\documentclass[onecolumn]{article}
\usepackage{ijcai21}

\usepackage{times}
\usepackage{soul}
\usepackage{url}
\usepackage[hidelinks]{hyperref}
\usepackage[utf8]{inputenc}
\usepackage[small]{caption}
\usepackage{graphicx}
\usepackage{subfigure}
\usepackage{amsmath}
\usepackage{bm}
\usepackage{mathtools}
\usepackage{amssymb}
\usepackage{amsthm}
\usepackage{booktabs}
\usepackage{algorithm}
\usepackage{algorithmic}
\usepackage{color}
\usepackage{multicol}
\usepackage{multirow}
\newtheorem{example}{Example}

\newtheorem{mydefine}{Definition}
\newtheorem{mypropt}{Proposition}
\newtheorem{myremark}{Remark}

\DeclareSymbolFont{symbolsC}{U}{txsyc}{m}{n}
\SetSymbolFont{symbolsC}{bold}{U}{txsyc}{bx}{n}
\DeclareMathSymbol{\Diamonddot}{\mathord}{symbolsC}{144}

\title{Neural Network for Weighted Signal Temporal Logic}

\author{Ruixuan Yan, and Agung Julius}

\begin{document}
	
	\maketitle
	
	\begin{abstract}  
		In this paper, we propose a  neuro-symbolic framework called weighted Signal Temporal Logic Neural Network (wSTL-NN) that combines the characteristics of neural networks and temporal logics. Weighted Signal Temporal Logic (wSTL) formulas are recursively composed of subformulas that are combined using logical and temporal operators. The quantitative semantics of wSTL is defined such that the quantitative satisfaction of subformulas with higher weights has more influence on the quantitative satisfaction of the overall wSTL formula. In the wSTL-NN, each neuron corresponds to a wSTL subformula, and its output corresponds to the quantitative satisfaction of the formula. We use wSTL-NN to represent wSTL formulas as features to classify time series data. STL features are more explainable than those used in classical methods. The wSTL-NN is end-to-end differentiable, which allows learning of wSTL formulas to be done using back-propagation. To reduce the number of weights, we introduce two techniques to sparsify the wSTL-NN. We apply our framework to an occupancy detection time-series dataset to learn a classifier that predicts the occupancy status of an office room.

	\end{abstract}
	
	\section{Introduction}
	
	Time series classification (TSC) has been considered as one of the most challenging tasks in machine learning (ML). Many supervised ML algorithms have been proposed to solve TSC problems \cite{qian2020dynamic,fawaz2019deep}. However, these classification models are often not human-readable, for example, hyperplanes in a higher-dimensional space. Temporal logics are formal languages that can express specifications about the temporal properties of systems. Compared with traditional ML models, temporal logic formulas can express temporal and logical properties in a human-readable form. Human-readability is important because it can give non-expert users insights into the model from qualitative and quantitative perspectives. Temporal logics have been widely used to solve formal verification and controller synthesis problems of cyber-physical systems~\cite{ijcai2020-570,rodrigues2020automatic,kantaros2020stylus,lindemann2020barrier,gundana2020event}. 
	
	Signal Temporal Logic (STL), a branch of temporal logic, is used to express specifications over real-time and real-valued data. STL has been a popular tool for analyzing real-time series data by learning STL formulas from data. The learning task is based on a notion of  quantitative satisfaction of STL formulas and posed as an optimization problem with the quantitative satisfaction in the objective function \cite{xu2019transfer,kong2014temporal}. A notion of quantitative satisfaction of STL was proposed in \cite{donze2010robust} that uses $\max$ and $\min$ functions, which are non-convex and non-smooth. As pointed out in \cite{mehdipour2020specifying}, the traditional quantitative satisfaction has a limitation that the satisfaction of parts of the formula other than the most "significant" part has no contribution to the overall quantitative satisfaction. Mehdipour et al then proposed a weighted STL (wSTL) that allows the expression of user preference on STL specifications by encoding the preference of satisfaction with weights. However, the quantitative satisfaction proposed in \cite{mehdipour2020specifying} has two limitations: 1) the subformulas with zero weights are still influential in the overall quantitative satisfaction; 2) it does not reflect that quantitative satisfaction of subformulas with higher weights has a greater influence on the overall quantitative satisfaction. 
	
	Combining neural networks and symbolic logic to perform learning tasks has attracted lots of attention \cite{riegel2020logical,yang2017differentiable,serafini2016logic,rocktaschel2017end} in recent years. Activation functions corresponding to truth functions of logical operators in first-order logic have been proposed in \cite{riegel2020logical} such that the truth value bounds of formulas can be learned from the neural network. Most of the existing STL learning algorithms solve a non-convex optimization problem to find the parameters in the formula \cite{kong2014temporal,xu2019information,yan2019swarm}, where the loss function is not differentiable everywhere with respect to the parameters. 
	
	The problem of using neural networks to perform temporal logic learning tasks has not been well studied. The contributions of this paper are: 1) we define a novel quantitative satisfaction for wSTL that has the properties of non-influence of zero weights, ordering of influence, and monotonicity (see Section {\ref{sec4}} for more details); 2) we construct a novel framework that combines the characteristics of neural network and wSTL such that it can perform wSTL learning tasks, where neurons represent the quantitative satisfaction of temporal and logical operators; 3) we propose two sparsification techniques that can prune the weights and reduce the memory. 
	
	\section{Preliminaries}
	
	A discrete $l$-dimensional time series data is denoted as $s = s(0),s(1),...$, where $l\in\mathbb{Z}_{>0}$, $s(k)\in\mathbb{R}^l$, $k\in\mathbb{Z}_{\geq 0}$. A time interval between $k_1$ and $k_2$ is denoted as $I = [k_1,k_2] = \{k'|k_1\leq k'\leq k_2,k_1,k_2\in\mathbb{Z}_{\geq0}\}$, and $k+I$ denotes the time interval $[k+k_1,k+k_2]$. 
	
	\subsubsection{Signal Temporal Logic (STL)}
	
	In this paper, we consider a fragment of Signal Temporal Logic (STL) proposed in \cite{maler2004monitoring}, whose syntax is defined recursively as follows:
	\begin{equation}
	\phi :=\top|\pi|\neg \phi|\phi_1 \wedge \phi_2|\phi_1 \vee \phi_2|\square_{I}\phi|\lozenge_{I}\phi,
	\end{equation}
	where $\phi,\phi_1,\phi_2$ are STL formulas, $\top$ is Boolean \textit{True}, $\pi:= f(s) =\bm{a}^Ts\leq c$ is a predicate defined over $s$, $\bm{a}\in \mathbb{R}^l, c\in \mathbb{R}$, $\phi, \phi_1, \phi_2$ are STL formulas, $\neg, \wedge, \vee$ are logical \textit{negation, conjunction,} and \textit{disjunction} operators. Temporal operators $\square,\lozenge$ read as "\textit{always, eventually}", respectively. $I$ is a time interval and  $\square_{I}\phi$ is satisfied at $k$ if $\phi$ is satisfied at all $k'\in k+I$, $\lozenge_{I}\phi$ is satisfied at $k$ if $\phi$ is satisfied at some $k'\in k+I$. The Boolean semantics of STL measures whether $s$ satisfies $\phi$ at $k$ qualitatively, for example, $(s,k)\models \phi$ reads as "$s$ satisfies $\phi$ at $k$" and $(s,k)\not\models\phi$ reads as "$s$ violates $\phi$ at $k$". The quantitative semantics of STL measures the degree of satisfaction or violation of $\phi$ over $s$ at $k$. 
	
	\begin{mydefine}\label{mydef2}
		The quantitative semantics (satisfaction) of STL is defined as \cite{donze2010robust}:
		\begin{equation}
		\begin{aligned}
		r(s,\pi,k) &=c-\bm{a}^Ts(k),\\
		r(s,\neg \phi,k)&=-r(s,\phi,k),\\
		r(s,\phi_1\wedge \phi_2,k)&=\min(r(s,\phi_1,k),r(s,\phi_2,k)),\\
		r(s,\phi_1\vee\phi_2,k)&=\max(r(s,\phi_1,k),r(s,\phi_2,k)),\\
		r(s,\square_{I}\phi,k)&=\min_{k'\in[k+k_1,k+k_2]}r(s,\phi,k'),\\
		r(s,\lozenge_{I}\phi,k)&=\max_{k'\in{[k+k_1,k+k_2]}}r(s,\phi,k').\\
		\end{aligned}\label{Quant_semant}
		\end{equation}
	\end{mydefine}
	\noindent The quantitative satisfaction at $k=0$ is simplified as $r(s,\phi)$. 
	
	\section{Weighted Signal Temporal Logic (wSTL)} \label{sec3}
	
\begin{figure}
		\centering
		\includegraphics[width =1.5 in]{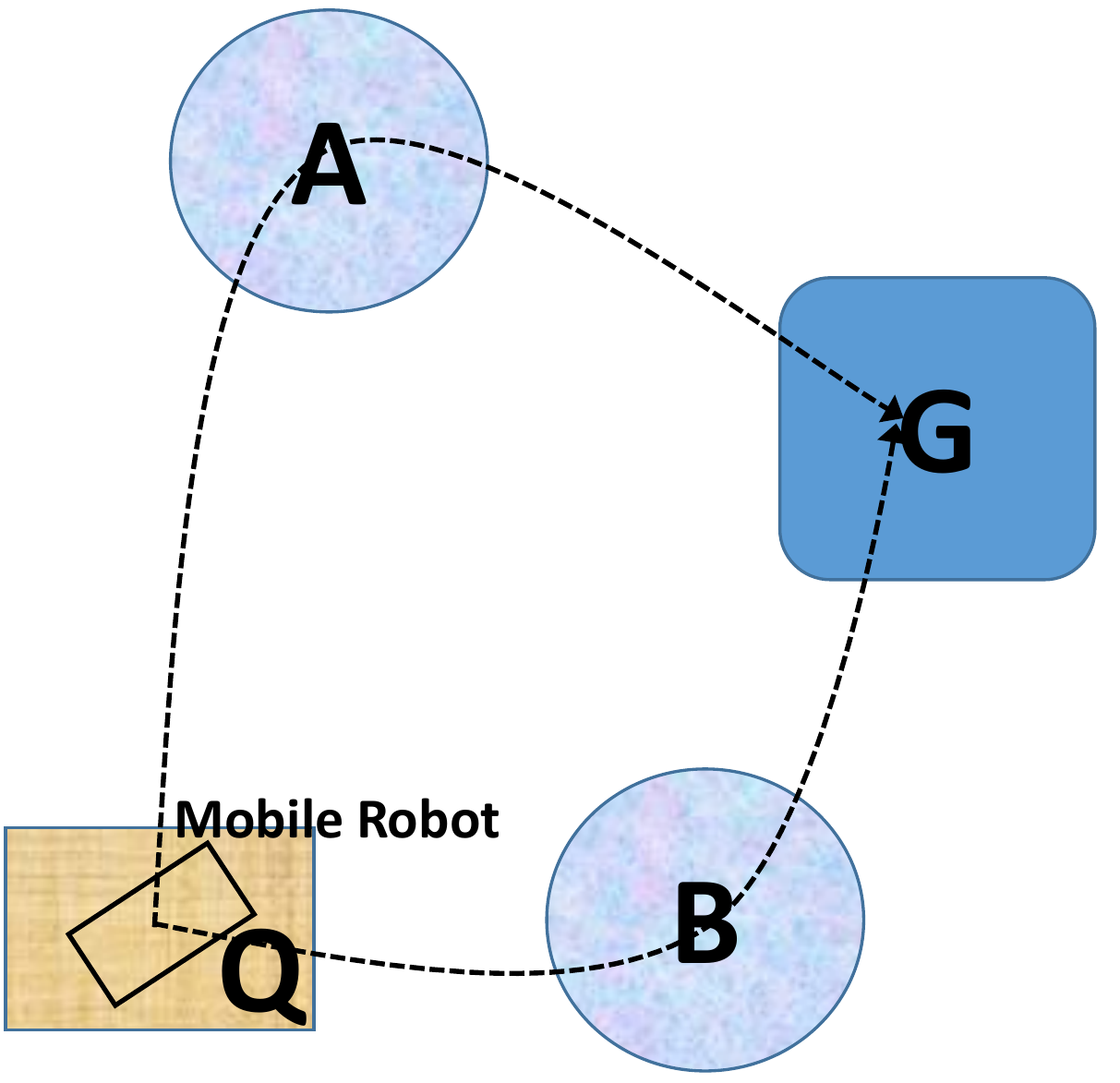}
		\caption{Illustration of a robot transporting building materials.}
		\label{figexp_buildmaterial}
	\end{figure}

	The quantitative satisfaction of the $\wedge,\square,\vee,\lozenge$ operators in \textbf{Definition \ref{mydef2}} is the minimum or maximum of the quantitative satisfaction of subformulas. This means the overall quantitative satisfaction is determined by the quantitative satisfaction of a single subformula. Also, traditional quantitative satisfaction cannot express importance over different subformulas. In many cases, we require the quantitative satisfaction function capable of reflecting the influence of different subformulas.
	\begin{example}
		A mobile robot needs to transport building materials from $Q$ to a construction site $G$ and it has to go to either store $A$ or $B$ for materials before arriving at $G$ (See Figure \ref{figexp_buildmaterial} as an illustration). We can use an STL formula $\phi= \lozenge_{[1,5]}(\pi_A \vee \pi_B) \wedge\lozenge_{[8,10]}\pi_G$ to express this requirement, where $\pi_A,\pi_B,\pi_G$ are predicates describing $A, B, G$, respectively. Formula $\phi$ reads as "The robot needs to arrive at $A$ or $B$ between 1 s and 5 s and arrive at $G$ between 8 s and 10 s." The problem of planning the movement of the robot is posed as an optimization problem by minimizing the energy and maximizing the quantitative satisfaction simultaneously. Even though the distance of going to $A$ is longer, it is favorable for the robot to go to store $A$ because the price is cheaper. In this way, the quantitative satisfaction of going $A$ should have a greater influence on the quantitative satisfaction of $\phi$ than the quantitative satisfaction of going $B$.
	\end{example}
	We introduce the notion of importance weights into STL, where quantitative satisfaction of subformulas with larger weights has a greater influence on the overall quantitative satisfaction. The novel STL is called weighted STL (wSTL).
	
	\begin{mydefine}
		The syntax of wSTL is defined over $s$ as
		\begin{equation}
		\tilde{\phi} := \top|\pi|\neg\tilde{\phi}|\prescript{w_1}{}{\tilde{\phi}_1} \wedge \prescript{w_2}{}{\tilde{\phi}_2}|\prescript{w_1}{}{\tilde{\phi}_1}\vee\prescript{w_2}{}{\tilde{\phi}_2}|\square_{I}^{\bm{w}}\tilde{\phi}| \lozenge_{I}^{\bm{w}}\tilde{\phi},
		\end{equation}
		where $\top,\pi$, and the logical and temporal operators are the same as the ones in STL, $w_1$ and $w_2$ are positive weights on the subformulas $\tilde{\phi}_1$ and $\tilde{\phi}_2$ correspondingly, $\bm{w} = [w_{k_1}, w_{k_1+1},...,w_{k_2}]^T\in \mathbb{R}_{>0}^{k_2-k_1+1}$ assigns a positive weight $w_{k'}$ to $k'\in[k_1, k_2]$ in the temporal operators.
	\end{mydefine}
	
	With importance weights, wSTL can express more complicated specifications. The Boolean semantics of wSTL formulas are the same as the corresponding STL formulas. Throughout the paper, we use $\bm{w}$ to denote weights associated with an operator and $\bm{w}^{\tilde{\phi}}$ to denote weights associated with a wSTL formula $\tilde{\phi}$ that may have multiple operators.
	
	\begin{mydefine}
		The quantitative semantics (satisfaction) of wSTL is defined as
		$$
		r^w(s,\pi,k) =c-\bm{a}^Ts(k),\nonumber
		$$	
		$$
		r^w(s,\neg \tilde{\phi},k)=-r^w(s,\tilde{\phi},k),\nonumber
		$$
		$$
		r^w(s,\prescript{w_1}{}{\tilde{\phi}_1}\wedge \prescript{w_2}{}{\tilde{\phi}_2},k) = \otimes ^\wedge([w_i, r^w(s,\tilde{\phi}_i, k)]_{i = 1,2}),
		$$
		$$r^w(s,\prescript{w_1}{}{\tilde{\phi}_1}\vee \prescript{w_2}{}{\tilde{\phi}_2},k) = \otimes ^\vee([w_i, r^w(s,\tilde{\phi}_i, k)]_{i = 1,2}),$$
		$$
		r^w(s,\square_{I}^{\bm{w}}\tilde{\phi},k)= \otimes^\square (\bm{w},[r^w(s,\tilde{\phi},k+k')]_{k'\in[k_1,k_2]}),\vspace{-15pt}
		$$
		\begin{equation}
		r^w(s,\lozenge_{I}^{\bm{w}}\tilde{\phi},k)= \otimes^\lozenge (\bm{w},[r^w(s,\tilde{\phi},k+k')]_{k'\in[k_1,k_2]}),
		\label{wSwarmSTLrbndef}
		\end{equation}		
		where $\otimes^\wedge,\otimes^\vee:\mathbb{R}_{>0}^2\times \mathbb{R}^2\rightarrow \mathbb{R}$,  $\otimes^\square,\otimes^\lozenge:\mathbb{R}_{>0}^{k_2-k_1+1}\times \mathbb{R}^{k_2-k_1+1}\rightarrow \mathbb{R}$ are activation functions corresponding to the $\wedge,\vee, \square,\lozenge$ operators, respectively. 
	\end{mydefine}
	
	\begin{myremark}
		The double negation property holds for the activation functions in (\ref{wSwarmSTLrbndef}) because $r^w(s,\neg\neg\tilde{\phi},k) = -r^w(s,\neg\tilde{\phi},k) = r^w(s,\tilde{\phi},k)$. 
	\end{myremark}
	
	\begin{myremark}
	The quantitative satisfaction in (\ref{wSwarmSTLrbndef}) satisfies Demorgan's law if and only if the activation functions for the $\wedge,\vee, \square,\lozenge$ operators satisfy Demorgan's law. For example, we can show that Demorgan's law holds for the $\wedge$ and $\vee$ operators, and the validity of Demorgan's law for the $\square$ and $\lozenge$ operators can be proved similarly. It follows that 
	$r(s, \neg(\prescript{w_1}{}{\neg\tilde{\phi}_1}\wedge \prescript{w_2}{}{\neg\tilde{\phi}_2)},k) =-\otimes ^\wedge([w_i, -r^w(s,\tilde{\phi}_i, k)]_{i = 1,2}).$
	As the aggregation function $\otimes^\wedge$ satisfies Demorgan's law, we have $-\otimes ^\wedge([w_i, -r^w(s,\tilde{\phi}_i, k)]_{i = 1,2}) = \otimes ^\vee([w_i,r^w(s,\tilde{\phi}_i, k)]_{i = 1,2})
	= r^w(s,\prescript{w_1}{}{\tilde{\phi}_1}\vee \prescript{w_1}{}{\tilde{\phi}_2},k).$
	\end{myremark}

	\section{Neural Network for wSTL (wSTL-NN)} \label{sec4}
	
	Structurally, a wSTL-NN is a graph composed of neurons representing signals and operators that are connected in the way determined by the formula. 
	Neurons representing predicates accept $s$ at single time points as inputs and have activation functions corresponding to $r^w(s,\pi, k)$, whose outputs are quantitative satisfaction of predicates. Neurons representing logical or temporal operators accept quantitative satisfaction of subformulas as inputs and have activation functions corresponding to the operators.  

	\subsection{Activation Functions for Logical and Temporal Operators}
	
	Gradient-based neural learning of importance weights $\bm{w}^{\tilde{\phi}}$ requires the activation functions to be differentiable with respect to $\bm{w}^{\tilde{\phi}}$, where $\bm{w}^{\tilde{\phi}}$ is all the weights in $\tilde{\phi}$. Activation functions for the $\vee,\square,\lozenge$ operators can be derived from the activation function for the $\wedge$ operator using DeMorgan's law. A weighted version of the quantitative satisfaction in (\ref{Quant_semant}) was proposed in \cite{mehdipour2020specifying}, where the activation function for the $\wedge$ operator is expressed as
	\begin{equation}
	\otimes ^\wedge([w_i, r_i]_{i = 1:N}) = \min_{i= 1:N}\{((\frac{1}{2}-\bar{w}_i)sign(r_i)+\frac{1}{2})r_i\},
	\label{actvfunmeh}
	\end{equation}
	where $r_i = r^w(s,\tilde{\phi}_i, k)$, and $\bar{w}_i = {w_i}/\sum_{j= 1}^N{w_j}$ is the normalized weight. The activation function in (\ref{actvfunmeh}) has two limitations: 1) If $w_i = 0$, then $r_i$ is still possible to be the one deciding the quantitative satisfaction, for example, if $N = 3$, $r_1 = 1,r_2=r_3=3$, and $\bar{w}_1 = 0, \bar{w}_2= 0.5, \bar{w}_3 = 0.5$, then the overall quantitative satisfaction is $\otimes^\wedge([w_i,r_i]_{i = 1:3}) = r_1$. This means even if the importance weight associated with $\tilde{\phi}_1$ is 0, it still can determine the overall quantitative satisfaction. 2) The quantitative satisfaction of $\tilde{\phi}$ is still determined by the signal at a single time point. To address these limitations, we propose several principles of the activation function for the $\wedge$ operator such that subformulas with higher weights have a greater influence on the overall quantitative satisfaction and the weights can be learned by a neuro-symbolic network. 
	
	The choice of the activation function for the $\wedge$ operator should obey the 
	following principles: 
	\begin{itemize}
		\item \textbf{Non-influence of zero weights:} $r^w(s,\tilde{\phi}_i,k)$ has no influence on $r^w(s,\prescript{w_1}{}{\tilde{\phi}_1}\wedge \prescript{w_2}{}{\tilde{\phi}_2},k)$ if $w_i = 0$;
		\item \textbf{Ordering of influence:} if $r^w(s,\tilde{\phi}_1,k) = r^w(s,\tilde{\phi}_2,k)$ and $w_1>w_2$, then we have
		$$
		\hspace{-10pt}\resizebox{.9\hsize}{!}{$
			\left|\frac{\partial r^w(s,\prescript{w_1}{}{\tilde{\phi}_1}\wedge \prescript{w_2}{}{\tilde{\phi}_2},k)}{\partial r^w(s,\tilde{\phi}_1,k)}\right|> \left|\frac{\partial r^w(s,\prescript{w_1}{}{\tilde{\phi}_1}\wedge \prescript{w_2}{}{\tilde{\phi}_2},k)}{\partial r^w(s,\tilde{\phi}_2,k)}\right|;$}
		\hspace{-8pt}
		$$
		\item \textbf{Monotonicity:}  $r^w(s,\prescript{w_1}{}{\tilde{\phi}_1}\wedge \prescript{w_2}{}{\tilde{\phi}_2},k)$ increases monotonically over $r^w(s,\tilde{\phi}_i,k)$, i.e. $\otimes^\wedge ([w_i, r^w(s,\tilde{\phi}_i,k)]_{i = 1,2})\leq \otimes^\wedge ([w_i, r^w(s,\tilde{\phi}_i,k)+d]_{i = 1,2})$, where $d\geq 0$.
	\end{itemize}
	This paper comes up with a concrete activation function for the $\wedge$ operator by introducing another variable $\sigma$, which promotes the activation functions for the other operators.
	\begin{mydefine}
		The activation functions for the $\wedge,\vee,\square,\lozenge$ operators in (\ref{wSwarmSTLrbndef}) are defined as 
		\begin{equation}
		\begin{split}
		&\otimes ^\wedge\!([w_i, r^w(s,\tilde{\phi}_i, k)]_{i = 1,2},\sigma)=\frac{\sum_{i=1}^2\bar{w}_is_ir_i}{\sum_{i=1}^2\bar{w}_is_i},\\
		&\otimes ^\vee([w_i, r^w(s,\tilde{\phi}_i, k)]_{i = 1,2},\sigma) =  -\frac{\sum_{i=1}^2\bar{w}_is_ir_i}{\sum_{i=1}^2\bar{w}_is_i},\\
		&\otimes^\square (\bm{w},[r^w(s,\tilde{\phi},k+i)]_{i\in k+I},\sigma) \!=\!\frac{\sum_{i=k_1}^{k_2}\bar{w}_is_ir_i}{\sum_{i=k_1}^{k_2}\bar{w}_is_i},\\
		&\otimes^\lozenge\!\! (\bm{w},[r^w(s,\tilde{\phi},k+i)]_{i\in k+I},\sigma)=\frac{\sum_{i=k_1}^{k_2}\bar{w}_is_ir_i}{-\sum_{i=k_1}^{k_2}\bar{w}_is_i},\\
		\end{split}
		\label{wSwarmSTL_actfun}
		\end{equation} 
		where $\bar{w}_i$ is the normalized weight ($\bar{w}_i = w_i/\sum_{j=1}^2{w_j}$ in the $\wedge,\vee$ operators, $\bar{w}_i = {w_i}/\sum_{j=k_1}^{k_2}{w_j}$ in the $\square, \lozenge$ operators), $I = [k_1,k_2]$, $r_i = r^w(s, \tilde{\phi}_i, k)$ in the $\wedge$ operator, $r_i = -r^w(s,\tilde{\phi}_i,k)$ in the $\vee$ operator, $r_i = r^w(s,\tilde{\phi},k+i)$ in the $\square$ operator, $r_i = -r^w(s,\tilde{\phi},k+i)$ in the $\lozenge$ operator, $s_i = e^{-\frac{r_i}{\sigma}}/\sum_{j=1}^2e^{-\frac{r_j}{\sigma}}$ in the $\wedge$ and $\vee$ operators, $s_i = e^{-\frac{r_i}{\sigma}}/{\sum_{j=k_1}^{k_2}e^{-\frac{r_j}{\sigma}}}$ in the $\square,\lozenge$ operators, $\sigma>0$. To clarify the notation, we use $\bm{w}$ to denote weights of an operator before normalization, and $\bm{\bar{w}}$ to denote weights of the same operator after normalization.
	\end{mydefine}
	\begin{mypropt}
		The activation functions defined in (\ref{wSwarmSTL_actfun}) satisfy the principles of non-influence of zero weights, ordering of influence, and monotonicity.
	\end{mypropt}

	\begin{myremark}
		For sufficiently small $\sigma$ and $w_1 = w_2$, $r^w(s,\prescript{w_1}{}{\tilde{\phi_1}}\!\wedge\! \prescript{w_2}{}{\tilde{\phi_2}},k)$ can be arbitrarily close to the traditional STL quantitative satisfaction, i.e.  $r^w(s,\prescript{w_1}{}{\tilde{\phi_1}}\!\wedge\! \prescript{w_2}{}{\tilde{\phi_2}},k) = \min\{
		r^w(s,{\tilde{\phi_1}},k),r^w(s,{\tilde{\phi_2}},k)\}$. 
	\end{myremark}
	\begin{figure*}
		\centering
		\includegraphics[width = 6in]{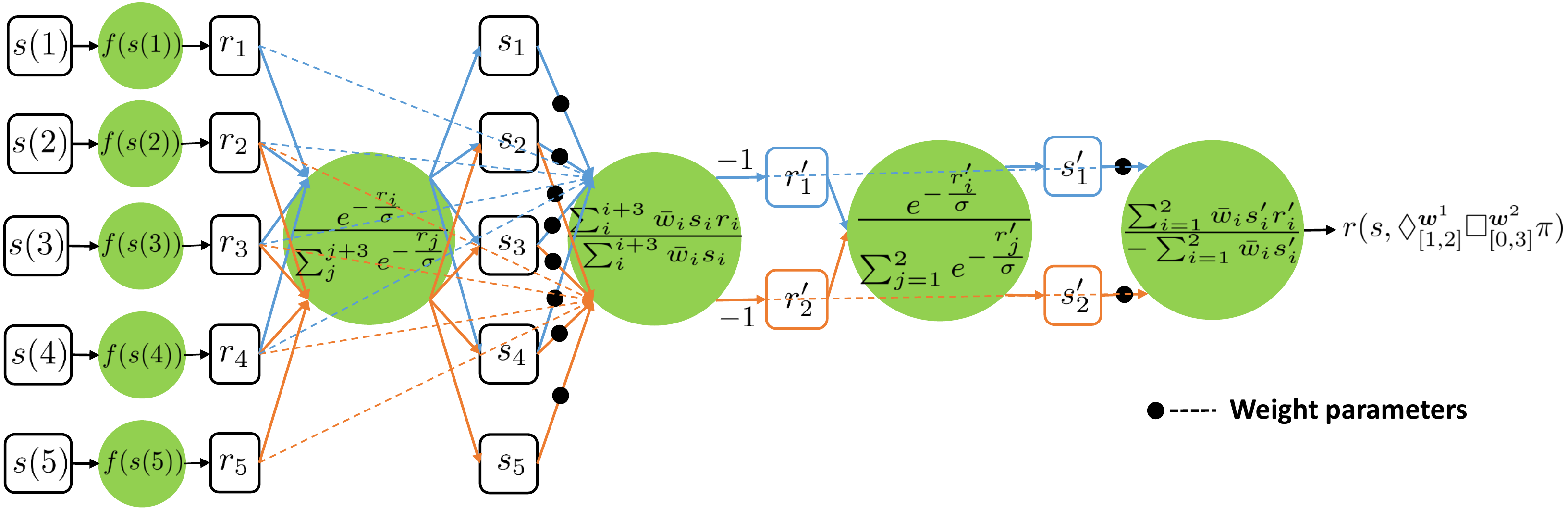}
		\caption{Neural network structure for $\tilde{\phi} = \lozenge_{[1,2]}^{\bm{w}^1}\square_{[0,3]}^{\bm{w}^2}\pi$.}
		\label{SwarmSTLNN_exp4}
	\end{figure*}
	
	With the activation function defined in (\ref{wSwarmSTL_actfun}), we can design neural networks for wSTL formulas. For example, the wSTL-NN for $\tilde{\phi} = \lozenge_{[1,2]}^{\bm{w}^1}\square_{[0,3]}^{\bm{w}^2}\pi$ is shown in Figure \ref{SwarmSTLNN_exp4}.
	
	\subsection{Learning of wSTL Formulas with wSTL-NN}
	
	In this paper, wSTL learning refers to the process of finding a wSTL formula that can classify two sets of time series data from a system. The two sets of data refer to a set ($D_P$) of positive data with label $1$ and a set ($D_N$) of negative data with label $-1$. The inferred formula is satisfied by the data in $D_P$ and violated by the data in $D_N$. 
	
	Suppose $|D_P| = \tilde{p}, |D_N| = \tilde{n}$, and $D_C = D_P\cup D_N$ that has $M = \tilde{p}+\tilde{n}$ data. The loss function should satisfy the requirements that the loss is small when the inferred formula is satisfied by the positive data or violated by the negative data, and the loss is large when these two conditions are not satisfied. 
	A candidate loss function that satisfies the above requirements could be \cite{yan2019swarm}
	\begin{equation}
	J(\tilde{\phi}) = \sum_{j = 1}^{M} h(l_j, r^w(D_C^j, \tilde{\phi})),
	\end{equation}
	where $D_C^j$ is the $j$-th data in $D_C$, $l_j$ is the label of $D_C^j$, and
	\begin{equation}
	\!h(l_j, r^w\!(D_C^j, \tilde{\phi}))\! =\! 
	\begin{cases}
	\zeta l_j r^w(D_C^j, \tilde{\phi}) & \!\text{if $l_j r^w(D_C^j, \tilde{\phi})\!>\!0$,}\hspace{-0.5cm}\\
	\gamma & \!\text{else,}
	\end{cases}
	\label{discreteloss}\\
	\end{equation}
	where $\zeta>0$ is a tuning parameter, and $\gamma$ is a positive large number that penalizes the cases when $\tilde{\phi}$ is violated by positive data or satisfied by negative data. However, this loss function is not differentiable everywhere with respect to $\bm{w}^{\tilde{\phi}},\bm{a}$, and $c$. An alternative loss function that is differentiable could be
	\begin{equation}
	J(\tilde{\phi}) = \sum_{j = 1}^M \exp(-\zeta l_jr^w(D_C^j, \tilde{\phi})).
	\label{contiloss}
	\end{equation}
	$J(\tilde{\phi})$ is small when $l_jr^w(D_C^j, \tilde{\phi})>0$ and grows exponentially when $l_jr^w(D_C^j, \tilde{\phi})<0$. Notice that the loss is proportional to $l_jr^w(D_C^j,\tilde{\phi})$ in (\ref{discreteloss}) and inverse proportional to $l_jr^w(D_C^j,\tilde{\phi})$ in (\ref{contiloss}), however, the loss in (\ref{contiloss}) still works because in the binary classification problem, positive and negative data have contradictory effects. We use Pytorch to build the wSTL-NN and perform the back-propagation to learn the formula. The wSTL learning algorithm with wSTL-NN is shown in \textbf{Algorithm \ref{wSTLinferAlg}}.	
	\begin{algorithm}[tb] 
		\caption{wSTL Learning Algorithm}
		\label{infer_alg}
		\textbf{Input}: Time-series data $D_C$, wSTL formula with specified structure $\tilde{\phi}$, number of iterations $K$, number of subformulas $\mathcal{J}$\\
		\textbf{Output}: Learned wSTL formula $\tilde{\phi}$
		
		\begin{algorithmic}[1] 
			\STATE Construct a wSTL-NN based on the structure of $\tilde{\phi}$, and initialize $\bm{w}^{\tilde{\phi}},\bm{a},c$.
			\FOR {$k = 1, 2, ..., K$}
			\STATE Select a mini-batch data $D_C^k$ from $D_C$.
			\FOR {$j = 1,2, ...,J$}
			\STATE Perform forward-propagation to compute the quantitative satisfaction of the $j$-th subformla using (\ref{wSwarmSTL_actfun}).
			\ENDFOR
			\STATE Compute the loss at the current iteration using (\ref{contiloss}).
			\STATE Perform back-propagation to update the parameters in the wSTL-NN.
			\ENDFOR
			\STATE \textbf{return} $\tilde{\phi}$
		\end{algorithmic}
		\label{wSTLinferAlg}
	\end{algorithm}

	\section{Sparsification of wSTL-NN} \label{sec5}
	
	As the structures of wSTL formulas become complex, the number of parameters to learn in the wSTL-NN will be large. As a result, the memory needed for back-propagation will be intensive. Sparsifying the weights of the wSTL-NN is one way to reduce the memory footprint of the network. 
	\begin{example}
		Consider we want to infer a formula to describe a mobile robot performing a package delivery task. Suppose the formula that we learn is $\tilde{\phi} = \square_{[0,4]}^{\bm{w}}\pi$, and the normalized weight is $\bar{\bm{w}} = [10^{-4},10^{-3},5\times 10^{-4},0.5, 0.4984]$ which indicates the data at time point $k = 3$ and $k = 4$ have greater influence on deciding whether a package delivery task is performed than the data at time points $k = 0,1,2$. Therefore, for this case, we could constrain the first three weights to $0$ and learn the last two weights, which reduces the memory required. Furthermore, if we want to use the inferred formula to monitor whether the robot has delivered a package, we could just check the satisfaction at $k=3$ and $k=4$. With weight sparsification techniques, we can set many weights that are less important to 0.
	\end{example}
	\subsection{Sparsification with Weight Thresholding} \label{sec51}
	
	During the back-propagation process, we can add additional constraints on the weights to set the less important weights to zero. Specifically, there are two perspectives to add constraints. The first one is by adding a threshold on the weights:
	\begin{equation}
	\tilde{w}_i = 
	\begin{cases}
	0 & \text{if $\bar{w}_i\leq \tau$},\\
	\bar{w}_i & \text{otherwise,}
	\end{cases}
	\label{thresholdtau}
	\end{equation}
	where $\tau$ is a specified threshold, and $\tilde{w}_i$ is the $i$-th weight after sparsification. 
	The second perspective is by thresholding the number ($\bar{s}$) of weights to keep, which is called top-$\bar{s}$ sparsification, i.e.
	\begin{equation}
	\tilde{w}_i = 
	\begin{cases}
	\bar{w}_i & \text{if $\bar{w}_i$ is one of the top-$\bar{s}$ weights},\\
	0 & \text{otherwise.}
	\end{cases}
	\label{thresholdkbar}
	\end{equation}
	
	\subsubsection{Weight Thresholding Analysis}
	
	In this part, we will analyze how much of the weights can be set to 0 such that the classification accuracy is not affected. For simplicity, we analyze the $\square$ operator, and the results of other operators are also straightforward to obtain. In general, $r^w(s,\square_I^{\bm{w}},k)$ can be written as
	\begin{equation}
	r^w(s,\square_{I}^{\bm{w}},k) = \sum_{i=1}^{I_+} \bar{w}_i^+ z_i^+ + \sum_{i=1}^{I_-} \bar{w}_i^-z_i^-,
	\end{equation}
	where $I_+(I_-)$ is the number of $r_i$ that is positive (negative),
	$$
	\resizebox{.99\hsize}{!}{$
		z_i^+ = \frac{e^{-r_i}r_i}{\sum_{j = 1}^{I_{+}+I_{-}}e^{-r_j}} (r_i>0), z_i^- = \frac{e^{-r_i}r_i}{\sum_{j = 1}^{I_++I_-}e^{-r_j}}(r_i<0),$} 
	$$
	and $\bar{w}_i^+ (\bar{w}_i^-)$ is the weight associated with $z_i^+ (z_i^-)$. If we want to keep the same classification accuracy, then the sign of the quantitative satisfaction should not change. We analyze how much $\bar{w}_i$ can change from the following perspectives:
	\begin{itemize}
		\item The quantitative satisfaction $r^w(s,\square_{I}^{\bm{w}},k)>0$, which means
		\begin{equation}
		\sum_{i = 1}^{I_+}\bar{w}_i^+ z_i^+ + \sum_{i = 1}^{I_-}\bar{w}_i^-z_i^->0. 
		\end{equation}
		If we set $\bar{w}_i^-=0$, then the quantitative satisfaction will not change sign. Hence we need to compute how much $\bar{w}_i^+$ can change such that $r^w(s,\square_{I}^{\bm{w}},k)$ will not change sign. Suppose $\min\{z_i^-\} = \delta$, $\max\{z_i^+\} = \gamma$ , $\bar{w}_s^+ = \sum_{i=1}^{I_+}\bar{w}_i^+$, and the fraction of $\bar{w}_s^+$ set to 0 is $\bar{f}$, we have
		\begin{equation}
		\begin{split}
		\gamma(1-\bar{f})\bar{w}_s^++\delta(1-\bar{w}_s^+)=0,\\
		\bar{f} = 1+\frac{\delta(1-\bar{w}_s^+)}{\gamma \bar{w}_s^+}.
		\end{split}
		\end{equation}
		\item The quantitative satisfaction $r^w(s,\square_{I}^{\bm{w}},k)<0$, which means
		\begin{equation}
		\sum_{i = 1}^{I_+}\bar{w}_i^+ z_i^+ + \sum_{i = 1}^{I_-}\bar{w}_i^-z_i^-<0. 
		\end{equation}
		Using the similar approach as above, we have
		\begin{equation}
		\begin{split}
		\delta(1-\bar{f})\bar{w}_s^-+\gamma(1-\bar{w}_s^-)=0,\\
		\bar{f} = 1+\frac{\gamma(1-\bar{w}_s^-)}{\delta \bar{w}_s^-},
		\end{split}
		\end{equation}
		where $\bar{w}_s^- = \sum_{i=1}^{I_-}\bar{w}_i^-.$
	\end{itemize}
	
	\subsection{Sparsification with Gate Variables}
	
	Apart from thresholding the weights, weight sparsification can also be achieved by introducing another variable called the gate variable. 
	\subsubsection{Gate Variables}
	
	Assume each weight variable $\bar{w}^{\tilde{\phi}}_i$ is associated with a sparsification variable $g_i^s\in\{0,1\}$, then the sparsified weight is $\tilde{w}^{\tilde{\phi}}_i = \bar{w}^{\tilde{\phi}}_ig_i^s$, where $\bar{w}^{\tilde{\phi}}_i$ is the $i$-th weight in $\tilde{\phi}$. This sparsification technique results in many weights being 0 if many $g_i^s$ are zero. To learn $g_i^s$, we adopt the method in \cite{srinivas2017training} that considers $g_i^s$ as a sample drawn from a Bernoulli random variable $g_i$. Hence we can learn the gate variables $g_i$ instead of $g_i^s$. To clarify the notation, we use $\bm{g}$ to denote the real-valued gate variables, and $\bm{g}^s$ to denote samples drawn from the Bernoulli distribution specified by $\bm{g}$. The learning process becomes learning the weight variables, variables in predicates, and gate variables simultaneously. 
	\subsubsection{Sparsification Realization}
	A critical point to obtain sparse weights is that the gate variables $\bm{g}$ are small, i.e., most $g_i$ is less than $0.5$. To facilitate this, we incorporate a so-called bi-modal regularizer that ensures most gate variables are small while only few gate variables are large \cite{murray2010algorithm}. The bi-modal regularizer is expressed as $g_i(1-g_i)$. Also, we incorporate the $\mathcal{L}_1$ regularizer into the loss function, which leads to
	$$
	J(\tilde{\phi},\bm{g})\! =\!\! \sum_{j = 1}^M\!\exp(-\zeta l_j r(D_C^j, \tilde{\phi})) + \lambda_1\! \!\sum_{i = 1}^m \!g_i (1-g_i) + \lambda_2 \!\sum_{i=1}^{m}\! g_i,
	$$
	where $m$ is the number of weights in $\tilde{\phi}$.

	\section{Case Study}
	
	In this section, we use an experiment to evaluate the performance of the wSTL learning algorithm. The dataset we use is the occupancy detection dataset from the UCI Machine Learning Repository \cite{Dua:2019,candanedo2016accurate}. The dataset is a time-series dataset that records the information including temperature ($s_1$), relative humidity ($s_2$), light intensity ($s_3$), CO$_2$ concentration ($s_4$), humidity radio ($s_5$), and occupancy of an office room from 2015-02-02 to 2015-02-18. The data is recorded every minute and the target variable is the occupancy (-1 for not occupied and 1 for occupied). 
	
	\subsection*{Learning wSTL Formulas without Sparsification}
	Since the dataset was published in 2016, many ML algorithms have been applied to this dataset to predict the occupancy status. These ML models are less interpretable and readable than the wSTL-NN classifier. The weights in these ML models represent how much each feature in the model contributes to the prediction, however, the model cannot be written in a human-readable form. The wSTL-NN classifier can not only reflect the contribution of features, but also express the model as a wSTL formula that is human-readable. The learned wSTL formula can tell us what kind of temporal property determines the room is occupied or not occupied. 
	
	To facilitate the learning of wSTL formulas, we combine $K_I$ consecutive instances of the original data that have the same label as one instance of data in our experiment. This means each sample in $D_P$ is a trajectory segment  associated to a period of constant occupation for $K_I$ minutes. Conversely, each sample in $D_N$ is a trajectory segment associated to a period of constant vacancy for $K_I$ minutes. In the experiment, we set $K_I = 16, K = 10, \sigma = 1, \zeta = 1$, which results in 744 training data and 186 testing data. The formula structure is specified as $\tilde{\phi} = \square_{[0,15]}^{\bm{w}}\pi$. The classification accuracy of wSTL-NN and other ML models is shown in Table \ref{MLmethodsonoccupancy}.
	\begin{table}
		\centering
		\begin{tabular}{lr}
			\toprule
			Method & Accuracy (\%)\\
			\midrule
			Linear Discriminate Analysis (LDA) & 100\\
			K-Nearest Neighbors (KNN)& 100\\
			Logistic Regression & 100\\
			Supported Vector Machine (SVM) & 100\\
			Naive Bayes & 99.46\\
			Random Forest & 99.46\\
			Decision Tree & 98.38\\
			\bf{wSTL-NN (this paper)} & 99.46\\
			\bottomrule
		\end{tabular}
		\caption{Classification accuracy on the occupancy dataset.}
		\label{MLmethodsonoccupancy}
	\end{table}
	We also evaluate the performance of the wSTL-NN using the evaluation measures of sensitivity, specificity, positive predictive value (PPV), and negative predictive value (NPV) \cite{fawcett2006introduction} as these are standard measures to evaluate a binary classification model. The results of these evaluation measures are shown in Table \ref{wSTLNNsingleresult}. 
	The wSTL formula we learn is 
	\begin{equation}
		\tilde{\phi} = \square_{[0,15]}^{\bm{w}} \pi,
	\end{equation}
	where the predicate $\pi$ is learned as
	\begin{equation}
	\begin{split}
	\pi:= &(6.4\times 10^{-6} s_1 +5.5\times 10^{-7} s_2-0.0075s_3 \nonumber\\
	& -6.9\times 10^{-6}s_4 +0.02 s_5\leq -0.9854), 
	\end{split}
	\end{equation}
	and the normalized weight is learned as
	\begin{equation}
	\begin{split}
	\bar{\bm{w}} = &[0.0186,0.0768,0.0988,0.1019,0.0240,0.0593,\\
	&0.0298,0.0616,0.0887,0.0243,0.1172,0.0662,\\
	&0.0126,0.035,0.1154,0.0695].
	\end{split}
	\end{equation}
	Formula $\tilde{\phi}$ reads as ``$\pi$ is always satisfied between 0 minute and 15 minute with importance weights $\bar{\bm{w}}$ at the corresponding time points". Compared with the classical ML models that are hyperplanes in an $80 = 16\times 5$ dimensional space, the predicate $\pi$ in the wSTL formula $\tilde{\phi}$ is only in a 5-dimensional space and $\tilde{\phi}$ can be written in a natural-language form. This means the wSTL formula is more readable than the other ML models.
	\begin{table}
	\centering
		\begin{tabular}{lrr}
			\toprule
			Evaluation measure & wSTL-NN & Decision Tree \\
			\midrule
			Sensitivity & 1& 0.959\\
			Specificity & 0.991&1\\
			PPV & 0.986&1\\
			NPV & 1& 0.974\\
			\bottomrule
		\end{tabular}
		\caption{Performance evaluation measures for wSTL-NN.}
		\label{wSTLNNsingleresult}	
	\end{table}
	Apart from comparing with the ML models, we also compare the wSTL formula learned by the wSTL-NN with the traditional STL formula learned by solving a non-convex optimization problem \cite{yan2019swarm}. The traditional STL formula that we learn is
	\begin{equation}
	\begin{split}
	\phi = &\square_{[1,9]} (0.0032 s_1 +0.0044 s_2 - 0.0013 s_3 \\
	&-0.0003 s_4 +0.0057 s_5\leq -0.4096).
	\end{split}
	\end{equation}
	Compared with the learning of STL formulas, another advantage of the wSTL learning algorithm is that it accelerates the learning process. The learning of the wSTL formula takes 0.385 s, but the learning of the STL formula takes 269.7 s. The experiments in this paper are implemented on a Windows laptop with a 1.9GHz Intel i7-8665U CPU and an 8 GB RAM.

	\subsection*{Learning wSTL Formulas with Sparsification}
	In this subsection, we perform the learning of wSTL formulas with the sparsification techniques discussed in \textbf{Section \ref{sec5}}. To save space, we only show the results for the techniques of top-$\bar{s}$ sparsification and sparsification with gate variables. The performance of these two techniques is shown in Table \ref{sparperformcompr}. From Table \ref{sparperformcompr} we could see for the same sparsification level $\bar{s} = 4$, the accuracy for the top-$\bar{s}$ sparsification is lower than the sparsification with gate variables. In reality, if the sparsification level obtained from the sparsification with gate variables is not satisfied, we could use top-$\bar{s}$ sparsification to further prune the weights, which may lower the accuracy. 
	
	\begin{table}
		\centering
		\begin{tabular}{lrr}
			\toprule
			Sparsification technique & $\bar{s}$ & Accuracy\\
			\midrule
			\multirow{7}*{Top-$\bar{s}$ sparsification}& 2 & 98.38\\
			~ & 4 & 98.38\\
			~& 6 & 98.92\\
			~ & 8 & 99.46\\
			~ & 10 & 99.46\\
			~ & 12 & 99.46\\
			~ & 14 & 99.46\\
			\midrule
			Sparsification with gate variables & 4 & 98.92\\
			\bottomrule
		\end{tabular}
		\caption{Results of wSTL-NN with sparsification.}
		\label{sparperformcompr}
	\end{table}
	
	\section{Conclusion and Future Work}
	
	We propose an extension of STL that is called wSTL by incorporating weights into the specifications, where the proposed quantitative satisfaction has three properties: non-influence of zero weights, ordering of influence, and monotonicity. A novel framework combining the characteristics of neural networks and wSTL is presented, which is called wSTL-NN. Due to the differentiable property of the quantitative satisfaction of wSTL, the parameters of wSTL can be learned from the wSTL-NN and the task of learning wSTL formulas can be accomplished. Two sparsification techniques of wSTL-NN are given such that wSTL-NN will be sparse even if the wSTL formula has a complex structure. Future work will explore extending wSTL-NN to learn wSTL formulas without specified time intervals and more computationally efficient quantitative satisfaction functions. 
	
	\clearpage

	\bibliographystyle{named}
	\bibliography{ijcai21}

\begin{thebibliography}{}

\bibitem[\protect\citeauthoryear{Candanedo and
  Feldheim}{2016}]{candanedo2016accurate}
Luis~M Candanedo and V{\'e}ronique Feldheim.
\newblock Accurate occupancy detection of an office room from light,
  temperature, humidity and co2 measurements using statistical learning models.
\newblock {\em Energy and Buildings}, 112:28--39, 2016.

\bibitem[\protect\citeauthoryear{Carr \bgroup \em et al.\egroup
  }{2020}]{ijcai2020-570}
Steven Carr, Nils Jansen, and Ufuk Topcu.
\newblock Verifiable rnn-based policies for pomdps under temporal logic
  constraints.
\newblock In Christian Bessiere, editor, {\em Proceedings of the Twenty-Ninth
  International Joint Conference on Artificial Intelligence, {IJCAI-20}}, pages
  4121--4127, 7 2020.
\newblock Main track.

\bibitem[\protect\citeauthoryear{Donz{\'e} and Maler}{2010}]{donze2010robust}
Alexandre Donz{\'e} and Oded Maler.
\newblock Robust satisfaction of temporal logic over real-valued signals.
\newblock In {\em International Conference on Formal Modeling and Analysis of
  Timed Systems}, pages 92--106. Springer, 2010.

\bibitem[\protect\citeauthoryear{Dua and Graff}{2019}]{Dua:2019}
Dheeru Dua and Casey Graff.
\newblock {UCI} machine learning repository, 2019.

\bibitem[\protect\citeauthoryear{Fawaz \bgroup \em et al.\egroup
  }{2019}]{fawaz2019deep}
Hassan~Ismail Fawaz, Germain Forestier, Jonathan Weber, Lhassane Idoumghar, and
  Pierre-Alain Muller.
\newblock Deep learning for time series classification: a review.
\newblock {\em Data Mining and Knowledge Discovery}, 33(4):917--963, 2019.

\bibitem[\protect\citeauthoryear{Fawcett}{2006}]{fawcett2006introduction}
Tom Fawcett.
\newblock An introduction to roc analysis.
\newblock {\em Pattern recognition letters}, 27(8):861--874, 2006.

\bibitem[\protect\citeauthoryear{Gundana and
  Kress-Gazit}{2020}]{gundana2020event}
David Gundana and Hadas Kress-Gazit.
\newblock Event-based signal temporal logic synthesis for single and
  multi-robot tasks.
\newblock {\em arXiv preprint arXiv:2011.00370}, 2020.

\bibitem[\protect\citeauthoryear{Kantaros and
  Zavlanos}{2020}]{kantaros2020stylus}
Yiannis Kantaros and Michael~M Zavlanos.
\newblock Stylus*: A temporal logic optimal control synthesis algorithm for
  large-scale multi-robot systems.
\newblock {\em The International Journal of Robotics Research}, 39(7):812--836,
  2020.

\bibitem[\protect\citeauthoryear{Kong \bgroup \em et al.\egroup
  }{2014}]{kong2014temporal}
Zhaodan Kong, Austin Jones, Ana Medina~Ayala, Ebru Aydin~Gol, and Calin Belta.
\newblock Temporal logic inference for classification and prediction from data.
\newblock In {\em Proceedings of the 17th international conference on Hybrid
  systems: computation and control}, pages 273--282, 2014.

\bibitem[\protect\citeauthoryear{Lindemann and
  Dimarogonas}{2020}]{lindemann2020barrier}
Lars Lindemann and Dimos~V Dimarogonas.
\newblock Barrier function based collaborative control of multiple robots under
  signal temporal logic tasks.
\newblock {\em IEEE Transactions on Control of Network Systems},
  7(4):1916--1928, 2020.

\bibitem[\protect\citeauthoryear{Maler and
  Nickovic}{2004}]{maler2004monitoring}
Oded Maler and Dejan Nickovic.
\newblock Monitoring temporal properties of continuous signals.
\newblock In {\em Formal Techniques, Modelling and Analysis of Timed and
  Fault-Tolerant Systems}, pages 152--166. Springer, 2004.

\bibitem[\protect\citeauthoryear{Mehdipour \bgroup \em et al.\egroup
  }{2020}]{mehdipour2020specifying}
Noushin Mehdipour, Cristian-Ioan Vasile, and Calin Belta.
\newblock Specifying user preferences using weighted signal temporal logic.
\newblock {\em arXiv preprint arXiv:2010.00752}, 2020.

\bibitem[\protect\citeauthoryear{Murray and Ng}{2010}]{murray2010algorithm}
Walter Murray and Kien-Ming Ng.
\newblock An algorithm for nonlinear optimization problems with binary
  variables.
\newblock {\em Computational Optimization and Applications}, 47(2):257--288,
  2010.

\bibitem[\protect\citeauthoryear{Qian \bgroup \em et al.\egroup
  }{2020}]{qian2020dynamic}
Bin Qian, Yong Xiao, Zhenjing Zheng, Mi~Zhou, Wanqing Zhuang, Sen Li, and
  Qianli Ma.
\newblock Dynamic multi-scale convolutional neural network for time series
  classification.
\newblock {\em IEEE Access}, 8:109732--109746, 2020.

\bibitem[\protect\citeauthoryear{Riegel \bgroup \em et al.\egroup
  }{2020}]{riegel2020logical}
Ryan Riegel, Alexander Gray, Francois Luus, Naweed Khan, Ndivhuwo Makondo,
  Ismail~Yunus Akhalwaya, Haifeng Qian, Ronald Fagin, Francisco Barahona, Udit
  Sharma, et~al.
\newblock Logical neural networks.
\newblock {\em arXiv preprint arXiv:2006.13155}, 2020.

\bibitem[\protect\citeauthoryear{Rockt{\"a}schel and
  Riedel}{2017}]{rocktaschel2017end}
Tim Rockt{\"a}schel and Sebastian Riedel.
\newblock End-to-end differentiable proving.
\newblock In {\em Advances in Neural Information Processing Systems}, pages
  3788--3800, 2017.

\bibitem[\protect\citeauthoryear{Rodrigues~da Silva \bgroup \em et al.\egroup
  }{2020}]{rodrigues2020automatic}
Rafael Rodrigues~da Silva, Vince Kurtz, and Hai Lin.
\newblock Automatic trajectory synthesis for real-time temporal logic.
\newblock {\em arXiv e-prints}, pages arXiv--2009, 2020.

\bibitem[\protect\citeauthoryear{Serafini and Garcez}{2016}]{serafini2016logic}
Luciano Serafini and Artur~d'Avila Garcez.
\newblock Logic tensor networks: Deep learning and logical reasoning from data
  and knowledge.
\newblock {\em arXiv preprint arXiv:1606.04422}, 2016.

\bibitem[\protect\citeauthoryear{Srinivas \bgroup \em et al.\egroup
  }{2017}]{srinivas2017training}
Suraj Srinivas, Akshayvarun Subramanya, and R~Venkatesh~Babu.
\newblock Training sparse neural networks.
\newblock In {\em Proceedings of the IEEE Conference on Computer Vision and
  Pattern Recognition Workshops}, pages 138--145, 2017.

\bibitem[\protect\citeauthoryear{Xu and Topcu}{2019}]{xu2019transfer}
Zhe Xu and Ufuk Topcu.
\newblock Transfer of temporal logic formulas in reinforcement learning.
\newblock In {\em Proceedings of the 28th International Joint Conference on
  Artificial Intelligence}, pages 4010--4018. AAAI Press, 2019.

\bibitem[\protect\citeauthoryear{Xu \bgroup \em et al.\egroup
  }{2019}]{xu2019information}
Zhe Xu, Melkior Ornik, A~Agung Julius, and Ufuk Topcu.
\newblock Information-guided temporal logic inference with prior knowledge.
\newblock In {\em 2019 American Control Conference (ACC)}, pages 1891--1897.
  IEEE, 2019.

\bibitem[\protect\citeauthoryear{Yan \bgroup \em et al.\egroup
  }{2019}]{yan2019swarm}
Ruixuan Yan, Zhe Xu, and Agung Julius.
\newblock Swarm signal temporal logic inference for swarm behavior analysis.
\newblock {\em IEEE Robotics and Automation Letters}, 4(3):3021--3028, 2019.

\bibitem[\protect\citeauthoryear{Yang \bgroup \em et al.\egroup
  }{2017}]{yang2017differentiable}
Fan Yang, Zhilin Yang, and William~W Cohen.
\newblock Differentiable learning of logical rules for knowledge base
  reasoning.
\newblock In {\em Advances in Neural Information Processing Systems}, pages
  2319--2328, 2017.

\end{thebibliography}
	
\end{document}